\documentclass[11pt]{article}
\usepackage[margin=1in]{geometry}
\usepackage[round,authoryear]{natbib}
\usepackage{latexsym}
\usepackage[T1]{fontenc}
\usepackage[utf8]{inputenc}
\usepackage{booktabs}
\usepackage{graphicx}
\usepackage{amsmath}
\usepackage{amssymb}
\usepackage{longtable}
\PassOptionsToPackage{hyphens}{url}
\usepackage{url}
\renewenvironment{abstract}{\begin{center}\bfseries Abstract\end{center}\begin{quote}\small}{\end{quote}}
\makeatletter
\renewcommand{\paragraph}{\@startsection{paragraph}{4}{\z@}{1.2ex \@plus 0.5ex \@minus 0.3ex}{-0.6em}{\normalfont\normalsize\bfseries}}
\renewcommand{\maketitle}{\begin{center}{\LARGE\bfseries \@title \par}\vspace{1em}{\large \@author \par}\vspace{1.5em}\end{center}}
\makeatother
\title{What Should We Remove? Evaluating the Reliability and Costs of
       RLVR Data Decontamination}
\author{
  Esther Xin \\
  Independent Researcher \\
  \texttt{estherxin0011@gmail.com}\\
  {\small Code and data: \url{https://github.com/ethxin0011/rlvr_data_contamination}}
}

\begin{document}
\maketitle

\begin{abstract}
Reinforcement learning with verifiable rewards (RLVR) math corpora are
routinely screened for benchmark overlap before release, but the
\emph{decontamination policy itself}---which matching rule to trust, and
at what cost---is rarely evaluated. We audit sixteen open RLVR math
corpora (1{,}673{,}514 corpus-specific training items) against six
benchmarks (MATH-500, GSM8K, AIME24, AIME25, AIME26, AMC23; 1{,}949
items) and evaluate eight cleaning policies built on an exact, lexical,
and semantic candidate pipeline against an independently, blindly
dual-reviewed reference set: 594 training--benchmark pairs split into a
calibration partition and a held-out partition of unseen corpora and
benchmark items, plus a complete-candidate-edge review of 80 training
items (83 pairs) that measures item-level false removal. Reviewers agreed
on 89.4\% and 95.9\% of the two sets before adjudication. More
aggressive semantic cleaning is not more reliable: at a 0.80 cosine
threshold, 52 of 59 sampled removed items (88\%) were distinct problems
wrongly flagged as contamination, and no threshold in our preregistered
grid (0.80--0.98) meets the preregistered 90\% conservative-precision
target. The existing judge-adjudicated policy reaches 98.1\% resolved
precision and 98.2\% candidate-frame recall on held-out corpora, with
zero confirmed false removals among 6 sampled removals, one unresolved;
this small sample does not establish zero false-removal risk. Exact
matching has 100\% observed precision but recovers only 7\% of confirmed
held-out exposure within the candidate frame; adding lexical matching
raises candidate-frame recall to 30\% at 70\% precision. We
release the reviewed labels, per-policy removal decisions, and coverage
accounting so that RLVR data releases can report a decontamination
policy against measured reliability and cost rather than a single
overlap count.
\end{abstract}

\section{Introduction}
\label{sec:intro}

Training corpora for reinforcement learning with verifiable rewards
(RLVR) on mathematics have grown rapidly, both as standalone releases
and as ingredients folded into larger blends used for frontier
post-training \citep{guo2025deepseekr1,shao2024deepseekmath}. Because
these corpora are built by scraping, generating, or recombining
problems from many upstream sources, a now-standard concern is that some
training items are, in fact, copies or near-copies of items in the
evaluation benchmarks those models are later scored on. Recent work has
begun auditing this directly: \citet{huang2026atlas} trace 1.45 million
RLVR instances back to twenty atomic sources and curate a decontaminated
replacement corpus, and \citet{yang2023rethinking} show that simple
string-matching decontamination is bypassed by paraphrase and
translation, motivating a stronger LLM-based detector.

Both lines of work answer a version of \emph{``how much contamination is
there, and how do we find more of it''}. Neither asks a question that is
equally consequential for anyone who has to ship a decontaminated
corpus: \emph{once you have a detector, which concrete removal policy
should you actually apply, and what does applying it cost you in
legitimate training data?} A policy that maximizes recall by flagging
anything semantically similar to a benchmark problem will, in a domain
like competition mathematics where many corpora are built from a small
number of recurring templates (changed numbers, re-ordered constraints,
renamed variables), also flag a large number of training items that are
not contamination at all---they are simply mathematically related
problems. Reporting only a removal count, without independently
verifying what fraction of removed items were correctly removed, makes
it impossible for a downstream user to judge whether a released
``decontaminated'' corpus lost 0.1\% or 10\% of its legitimately useful
content.

We address this gap directly. Reusing an existing multi-tier
contamination-candidate pipeline (exact match, lexical $n$-gram/MinHash,
and embedding-based semantic retrieval, with an existing LLM-judge
adjudication layer) as our candidate generator, we define eight concrete,
fully specified cleaning policies and evaluate all of them against a new,
independently and blindly dual-annotated reference set that we construct
specifically for this purpose. Our reference set is split, before any
label is seen, into a calibration partition and a held-out partition
built from three corpora and benchmark questions that never appear in
calibration, so that we can test whether a policy tuned on one slice of
the data continues to behave reliably on genuinely unseen corpora. We
additionally draw a second sample of 80 candidate-bearing training items
and review \emph{every} candidate edge attached to each one, because a
single flagged pair is not sufficient evidence that an item was wrongly
removed: the same item may have another, genuinely contaminating, match.

\paragraph{Contributions.}
(i) A preregistered, independently and blindly dual-reviewed reference
set of 594 training--benchmark pairs plus 80 fully-resolved training
items (83 candidate edges), split into disjoint calibration and
held-out partitions by both benchmark question and training-example
cluster, to our knowledge the first such resource built specifically to
evaluate \emph{decontamination policies} rather than to characterize
contamination itself;
(ii) a item-level false-removal methodology that reviews all candidate
edges for a sampled item before counting it as wrongly removed, rather
than relying on single-pair precision, which we show would otherwise
overstate how conservative a policy is;
(iii) an empirical result that a common, simple cleaning heuristic
(semantic similarity above a fixed cosine threshold) trades reliability
for candidate-frame recall overall as its threshold is relaxed, with
the majority of its sampled removals at a permissive threshold confirmed
false on manual review; and
(iv) a full accounting of what each policy costs in retained corpus
size and source-composition shift, released alongside the reviewed
labels so other groups can audit, reuse, or extend this evaluation to
additional corpora, benchmarks, or policies.

\section{Related Work}
\label{sec:related}

\paragraph{RLVR data lineage and contamination auditing.}
\citet{huang2026atlas} is the closest prior work. ATLAS traces the
provenance of 1.45 million RLVR instances to twenty atomic sources,
documents that most released RLVR datasets are variants of a small
shared pool, and uses this lineage information together with a
source-level counterfactual utility score to curate a new, decontaminated
training set, DAPO++. Their contribution is a lineage-aware curation
method and the resulting dataset. Ours is different in kind: we do not
propose a new training corpus or a new detector. We take an already-built
detection pipeline as given and ask how \emph{reliable} a range of
concrete, fully specified removal rules built on top of it are, measured
against independently reviewed ground truth, including on corpora held
out from any tuning. ATLAS's analysis does not report a held-out
generalization check of a specific contamination threshold against
blinded human labels, nor an item-level false-removal rate; both are the
central quantities we report.

\paragraph{Decontamination method strength.}
\citet{yang2023rethinking} show that n-gram-based string matching misses
paraphrased and translated test-set variants, and propose a stronger
LLM-based decontamination method, demonstrating previously unknown
overlap in several widely used pretraining corpora. Their contribution
is a \emph{more sensitive detector}. We do not propose a new detector;
instead, we take detector outputs across several existing tiers (exact,
lexical, semantic, judge-adjudicated) as given and ask which concrete
decision rule built on those outputs is reliable enough to apply, and at
what data-loss cost, which their evaluation does not report.

\paragraph{Corpus provenance.}
One of the sixteen corpora we audit, \textsc{GooseReason-0.7M}, is
built by masking spans of real corpus text and generating plausible
distractors around them \citep{lu2026goldengoose}. We treat it here only
as one of sixteen audited math corpora, evaluated on the same
contamination-removal question as the other fifteen; we make no claim
about, and do not reuse any apparatus from, the separate question of
whether its generated-distractor construction introduces an
authenticity artifact exploitable by RLVR training, which is outside
this paper's scope.

\paragraph{Annotation reliability.}
Measuring reviewer agreement is necessary but not sufficient evidence
that reviewed labels are correct: raw percent agreement can overstate
reliability relative to chance-corrected statistics, and label quality
frameworks caution against treating any single annotation pass as ground
truth without further validation \citep{james2026consensus,
lavitas2021quality}. We report raw agreement rates for both of our
annotation passes and discuss this limitation explicitly in
\S\ref{sec:limitations} rather than treating our reviewed labels as an
infallible gold standard.

\section{Audited Corpora and Benchmarks}
\label{sec:data}

We audit sixteen open RLVR math training corpora, drawn from
widely used public releases spanning late 2024 through late 2025.
\begin{sloppypar}\small\texttt{acereason\_math, bigmath\_rl,
ctrl\_gsm8k\_train, ctrl\_math\_train, dapo\_math\_17k, deepmath\_103k,
deepscaler, dolci\_rlzero\_math, dolci\_think\_rl\_math,
goosereason\_math, nemotron\_rl\_math\_v2, nemotron\_rl\_omr,
numinamath15\_checkable, openr1\_math, polaris\_53k,
skywork\_or1\_math.}\end{sloppypar}
\noindent
We evaluate these corpora against six evaluation benchmarks: MATH-500
(500 items), GSM8K test (1{,}319), AIME24 (30), AIME25 (30), AIME26 (30,
released after the construction of most of the audited corpora), and
AMC23 (40). After normalization and corpus-specific deduplication, our
unit of analysis is a \emph{corpus-specific training item}: a unique
normalized problem string within one specific corpus, so that a problem
appearing in two corpora is counted, and can be independently removed or
retained, in each. This yields 1{,}673{,}514 corpus-specific training
items over 1{,}197{,}691 unique underlying problem texts.

\section{Method}
\label{sec:method}

\subsection{Existing candidate pipeline (reused, not a contribution)}
\label{sec:pipeline}

Every training item is compared against every benchmark item through an
existing three-tier matching pipeline, which we reuse unmodified as our
candidate generator. \textbf{Tier 1 (exact/near-exact)} normalizes
whitespace, casing, and LaTeX macros and checks for exact or
near-exact string equality. \textbf{Tier 2 (lexical)} computes
13-gram and 8-gram coverage overlap and a 128-permutation MinHash/LSH
Jaccard estimate between each training item and each benchmark item.
\textbf{Tier 3 (semantic)} embeds every training item and benchmark item
with Qwen3-Embedding-0.6B and retrieves nearest neighbors by cosine
similarity. A subset of tier-2/tier-3 candidate pairs was previously
adjudicated by an LLM judge into one of four labels (\textsc{same},
\textsc{subsumes}, \textsc{variant}, \textsc{different}); we refer to
the resulting rule---flag a pair if Tier 1 fires, or the judge labeled
it \textsc{same}/\textsc{subsumes}, or (for unjudged pairs) a fixed
lexical/semantic heuristic fires---as the \textbf{existing adjudicated
policy}. Full tier thresholds are given in Table~\ref{tab:config}.

\subsection{Policies under evaluation}
\label{sec:policies}

We evaluate eight fully specified decision rules for whether a training
item should be \emph{removed} because it matches a benchmark item:
\textbf{exact} (Tier 1 only); \textbf{lexical} (Tier 1, or
13-gram coverage $\geq 0.5$, or MinHash Jaccard $\geq 0.7$); five
\textbf{semantic-union} policies at cosine thresholds
$\{0.80, 0.85, 0.90, 0.95, 0.98\}$ (lexical policy, or semantic
cosine similarity at or above the stated threshold); and the
\textbf{existing adjudicated} policy described above. Every policy is a
strict superset rule built from the same underlying scores, so
differences between policies isolate the effect of the decision
threshold, not a difference in what evidence was available.

\subsection{Independent, blinded reference labels}
\label{sec:labels}

\paragraph{Pair-level sample.} We stratify the full candidate population
(21{,}796 corpus-specific candidate pairs) by corpus, benchmark, and a
joint stratum of (exact, lexical, adjudicated) policy flags crossed with
five semantic-score bands, and draw up to two pairs per stratum cell. We
partition benchmark questions 60/40 into calibration and a held-out set
\emph{before} sampling, assign three corpora (bigmath\_rl,
goosereason\_math, nemotron\_rl\_math\_v2) entirely to the held-out
partition, and then exclude from calibration any pair whose training
item shares a near-duplicate cluster with a held-out pair, or whose
benchmark question appears in the held-out partition, so that no
training-example family or benchmark question is shared between
partitions. This yields 594 sampled pairs: 448 calibration, 146
held-out.

Two reviewers independently labeled all 594 pairs, blind to every
policy flag, similarity score, corpus identity, and the opposing
reviewer's label, choosing one of six labels: \textsc{equivalent}
(same problem, possibly reworded), \textsc{solution\_exposure}
(benchmark-specific solution information present without being the same
problem), \textsc{changed\_instance} (same template or scenario, but
different numbers/constraints yielding a different answer),
\textsc{related\_distinct} (related topic or technique, different
problem), \textsc{unrelated}, or \textsc{uncertain}. Reviewers agreed on
531/594 pairs (89.4\%) before adjudication; a third pass independently
adjudicated all 63 disagreements using the same label set and the
original problem text, without access to any policy score.

\paragraph{Item-level sample.} Pair-level precision can overstate how
conservative a policy truly is: if an item has two candidate matches and
only one is reviewed, labeling the reviewed pair as a false positive does
not establish that removing the \emph{item} was a mistake, since the
other, unreviewed match might be genuine contamination. We therefore
drew a second, independent sample of five candidate-bearing training
items per corpus (80 items total, one corpus contributing all available
items where fewer than five existed) and exported \emph{every} candidate
edge for each sampled item (83 pairs). Nine of these 83 pairs had
identical problem/benchmark/answer text to an already-reviewed pair from
the first sample and reused that pair's adjudicated label directly,
without re-review; the remaining 74 were independently dual-reviewed
under the same protocol as above (71/74, 95.9\%, raw agreement; the
remaining 3 disagreements adjudicated). An item is counted as a
\emph{confirmed false removal} only if a policy removes it \emph{and}
every one of its reviewed candidate edges is labeled
\textsc{changed\_instance}, \textsc{related\_distinct}, or
\textsc{unrelated}; items with at least one \textsc{uncertain} edge and
no confirmed exposure are reported separately as \emph{unresolved}, not
forced into either category.

\subsection{Preregistered threshold-selection rule}
\label{sec:threshold}

To avoid selecting a semantic threshold by inspecting held-out
performance, we fixed, before reviewing any label, a candidate grid
$\{0.80,0.85,0.90,0.95,0.98\}$ and a target: the smallest (most
inclusive) threshold on the \emph{calibration} partition whose
conservative precision---treating every \textsc{uncertain} label as
incorrect---is at least 0.90. If no threshold on the grid meets this
target, we report that failure directly rather than relaxing the target
or selecting a threshold using held-out labels.

\subsection{Metrics}
\label{sec:metrics}

For a given partition and policy we report \emph{resolved precision}
(true positives over true-or-false positives, excluding
\textsc{uncertain} pairs from the denominator), a \emph{conservative
lower bound} that instead treats every \textsc{uncertain} pair as a
false positive, \emph{candidate-frame recall} (true positives over
true positives plus confirmed-exposure pairs the policy missed, within
our sampled candidate frame only, not the full corpus), and the
\emph{false-positive rate} among confirmed-negative pairs. All
pair-level statistics are inverse-probability weighted by each pair's
stratum sampling probability. Item-level statistics use the same
inverse-probability weighting over each sampled item's corpus-specific
sampling probability, with exploratory 95\% intervals computed by
resampling items within corpus 1{,}000 times; with five items per
corpus these intervals are reported as exploratory, not as rigorous
survey-design confidence intervals (\S\ref{sec:limitations}).

\section{Results}
\label{sec:results}

\begin{table*}[htbp]
\centering
\footnotesize
\setlength{\tabcolsep}{4.5pt}
\begin{tabular}{llrrrrr}
\toprule
Split & Policy & Precision & Prec.\ (lower) & Recall (cand.) & FPR & Weighted FP \\
\midrule
calibration & exact                & 1.0000 & 1.0000 & 0.0851 & 0.0000 & 0.0 \\
calibration & lexical              & 0.1763 & 0.1763 & 0.2596 & 0.0560 & 285.0 \\
calibration & existing adjudicated & 0.9546 & 0.8123 & 0.9851 & 0.0022 & 11.0 \\
calibration & semantic $\geq 0.80$ & 0.0511 & 0.0496 & 1.0000 & 0.8583 & 4366.5 \\
calibration & semantic $\geq 0.85$ & 0.1072 & 0.1050 & 0.9191 & 0.3535 & 1798.5 \\
calibration & semantic $\geq 0.90$ & 0.2167 & 0.2122 & 0.6681 & 0.1115 & 567.5 \\
calibration & semantic $\geq 0.95$ & 0.2091 & 0.2077 & 0.3915 & 0.0684 & 348.0 \\
calibration & semantic $\geq 0.98$ & 0.2083 & 0.2078 & 0.3191 & 0.0560 & 285.0 \\
\addlinespace
held-out & exact                & 1.0000 & 1.0000 & 0.0704 & 0.0000 & 0.0 \\
held-out & lexical              & 0.6975 & 0.6564 & 0.3005 & 0.0169 & 55.5 \\
held-out & existing adjudicated & 0.9812 & 0.9447 & 0.9824 & 0.0024 & 8.0 \\
held-out & semantic $\geq 0.80$ & 0.1596 & 0.1580 & 0.9906 & 0.6767 & 2222.5 \\
held-out & semantic $\geq 0.85$ & 0.2809 & 0.2783 & 0.9695 & 0.3220 & 1057.5 \\
held-out & semantic $\geq 0.90$ & 0.6661 & 0.6504 & 0.8779 & 0.0571 & 187.5 \\
held-out & semantic $\geq 0.95$ & 0.8153 & 0.7830 & 0.6268 & 0.0184 & 60.5 \\
held-out & semantic $\geq 0.98$ & 0.6975 & 0.6564 & 0.3005 & 0.0169 & 55.5 \\
\bottomrule
\end{tabular}
\caption{Pair-level policy metrics on the calibration partition (448
independently reviewed pairs) and the held-out partition (146 pairs, from
three corpora and benchmark questions never used in calibration). All
statistics are inverse-probability weighted by stratum sampling
probability. ``Prec.\ (lower)'' treats every \textsc{uncertain} label as
a false positive. ``Recall (cand.)'' is candidate-frame recall, not
whole-corpus recall. No semantic threshold on the preregistered grid
meets the 0.90 conservative-precision target on calibration
(\S\ref{sec:threshold}); the existing adjudicated policy exceeds it on
held-out only (0.9447), not on calibration (0.8123).}
\label{tab:pairmetrics}
\end{table*}

\begin{table}[htbp]
\centering
\small
\setlength{\tabcolsep}{5pt}
\begin{tabular}{lrrr}
\toprule
Policy & Removed & Confirmed false & Unresolved \\
\midrule
exact                & 0  & 0  & 0 \\
lexical              & 4  & 1  & 0 \\
existing adjudicated & 6  & 0  & 1 \\
semantic $\geq 0.80$ & 59 & 52 & 2 \\
semantic $\geq 0.85$ & 25 & 19 & 1 \\
semantic $\geq 0.90$ & 7  & 3  & 1 \\
semantic $\geq 0.95$ & 5  & 1  & 1 \\
semantic $\geq 0.98$ & 4  & 1  & 0 \\
\bottomrule
\end{tabular}
\caption{Item-level removal review: 80 candidate-bearing training items
(83 candidate edges; every edge for each sampled item was reviewed).
``Confirmed false'' counts items a policy removes for which every
reviewed candidate edge is labeled \textsc{changed\_instance},
\textsc{related\_distinct}, or \textsc{unrelated}. ``Unresolved'' counts
removed items with at least one \textsc{uncertain} edge and no confirmed
exposure. At threshold 0.80, 52 of 59 sampled removed items (88\%) are
confirmed false removals, not a demonstrated population-level rate.
The sampled false-removal rate generally falls with a higher threshold,
but rises from 20\% at 0.95 to 25\% at 0.98. Among policies that removed
sampled items, the existing adjudicated policy alone has zero confirmed
false removals (6 removals, one unresolved); this does not establish
zero false-removal risk.}
\label{tab:itemremoval}
\end{table}

\begin{table}[htbp]
\centering
\small
\setlength{\tabcolsep}{4pt}
\begin{tabular}{lrrrr}
\toprule
Policy & Removed & Removal \% & Retained & Source TVD \\
\midrule
exact                & 82     & 0.0049\% & 1{,}673{,}432 & 0.000026 \\
lexical              & 1{,}021  & 0.0610\% & 1{,}672{,}493 & 0.000133 \\
existing adjudicated & 1{,}314  & 0.0785\% & 1{,}672{,}200 & 0.000317 \\
semantic $\geq 0.80$ & 16{,}067 & 0.9601\% & 1{,}657{,}447 & 0.001146 \\
semantic $\geq 0.85$ & 7{,}701  & 0.4602\% & 1{,}665{,}813 & 0.000660 \\
semantic $\geq 0.90$ & 2{,}720  & 0.1625\% & 1{,}670{,}794 & 0.000318 \\
semantic $\geq 0.95$ & 1{,}491  & 0.0891\% & 1{,}672{,}023 & 0.000241 \\
semantic $\geq 0.98$ & 1{,}106  & 0.0661\% & 1{,}672{,}408 & 0.000157 \\
\bottomrule
\end{tabular}
\caption{Full-population removal cost across all 1{,}673{,}514
corpus-specific training items (not sampled). ``Source TVD'' is the
total variation distance between the per-corpus share of the pool
before and after removal, under 1.0 maximum distance. Threshold-0.80
semantic cleaning removes over twelve times as many items as the
existing adjudicated policy, while aggregate source-composition shift
remains small under every policy. Table~\ref{tab:itemremoval} separately
shows 52 confirmed false removals among 59 sampled removals at threshold
0.80; this does not establish a population-level false-removal rate.}
\label{tab:coverage}
\end{table}

\subsection{No semantic threshold meets the preregistered target}
\label{sec:result-threshold}

Table~\ref{tab:pairmetrics} reports pair-level metrics for all eight
policies on both partitions. On calibration, conservative precision for
every semantic-union threshold on our grid is below the 0.90 target
(ranging from 0.0496 at 0.80 to 0.2122 at 0.90), so, following
\S\ref{sec:threshold}, we select \emph{no} semantic threshold and report
this directly as a negative calibration result, rather than loosening
the target to accommodate the held-out partition.

\subsection{Aggressive semantic cleaning is less reliable, not more}
\label{sec:result-semantic}

The item-level review (Table~\ref{tab:itemremoval}) makes the practical
consequence concrete. At cosine threshold 0.80, the policy would remove
59 of our 80 sampled candidate-bearing items; of those, 52 (88\%) were
confirmed on manual review to be \textsc{changed\_instance},
\textsc{related\_distinct}, or \textsc{unrelated}---that is, confirmed
\emph{false} removals within our reviewed candidate frame. This 88\%
is a descriptive rate among sampled removals, not a demonstrated
population-level false-removal rate. The sampled false-removal rate
generally falls as the threshold is raised (19/25 at 0.85; 3/7 at 0.90;
1/5 at 0.95), but rises from 20\% at 0.95 to 25\% at 0.98 (1/4);
the improvement is therefore not monotonic. The existing adjudicated
policy removed 6 of the 80 items with zero confirmed false removals
(one unresolved case with an \textsc{uncertain} edge). These six
observations do not establish zero false-removal risk.
Held-out pair-level results in Table~\ref{tab:pairmetrics} corroborate
the overall trade-off: moving from threshold 0.80 to 0.95 trades candidate-frame recall
of 0.99 for 0.63 while precision rises from 0.16 to 0.82---there is no
threshold on our grid that achieves both candidate-frame recall and
resolved precision of at least 0.90 on the held-out partition.

\subsection{The existing adjudicated policy generalizes to held-out corpora}
\label{sec:result-generalization}

On the three held-out corpora and benchmark questions never used during
calibration, the existing adjudicated policy achieves 0.9812 resolved
precision (0.9447 conservative lower bound) and 0.9824 candidate-frame
recall. Resolved precision and candidate-frame recall are similar to
their calibration-partition values (0.9546 and 0.9851 respectively),
supporting reliability on these held-out corpora within the sampled
candidate frame. Conservative precision differs: it is 0.8123 on
calibration and 0.9447 on held-out, so the adjudicated policy exceeds
the 0.90 conservative-precision target on held-out only. Exact matching
has 100\% observed precision in both partitions but recovers only
7--9\% of confirmed exposure within the candidate frame; adding lexical
matching raises held-out candidate-frame recall to 30\% while resolved
precision falls to 70\%, below the existing adjudicated policy on both
axes.

\subsection{Data-loss cost scales with aggressiveness, but total corpus
composition shift stays small}
\label{sec:result-coverage}

Table~\ref{tab:coverage} reports full-population removal counts (not
sampled) across all 1{,}673{,}514 corpus-specific items. The existing
adjudicated policy removes 1{,}314 items (0.079\%); threshold-0.80
semantic cleaning removes 16{,}067 (0.96\%), over twelve times as many.
In the separate item-level sample, 52 of 59 removals under the latter
policy were confirmed false; this does not establish the fraction of
false removals among all 16{,}067 population removals.
Aggregate source-composition shift (total variation
distance between the corpus-share distribution before and after
removal) remains small even under the most aggressive policy
(0.0011), because removals are spread across corpora roughly in
proportion to their size. The demonstrated costs are therefore greater
full-population data removal and frequent confirmed false removals in
the reviewed sample, not a gross distortion of which corpora dominate
the cleaned pool.

\subsection{Summary figures}
\label{sec:result-figures}

\begin{figure}[htbp]
\centering
\includegraphics[width=\linewidth]{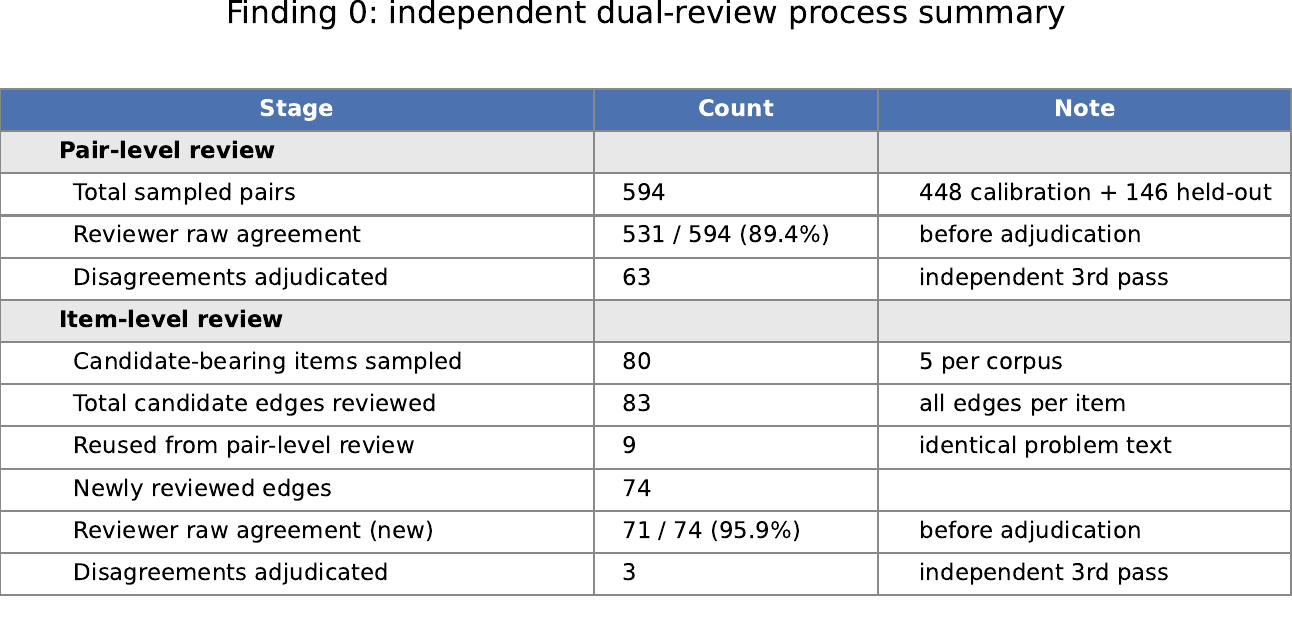}
\caption{Independent dual-review process summary: sample sizes and raw
reviewer agreement for both the pair-level sample (594 pairs, used for
Table~\ref{tab:pairmetrics}) and the item-level complete-candidate-edge
sample (83 pairs over 80 items, used for Table~\ref{tab:itemremoval}),
before and after adjudication of disagreements (\S\ref{sec:labels}).}
\label{fig:reviewsummary}
\end{figure}

\begin{figure}[htbp]
\centering
\includegraphics[width=\linewidth]{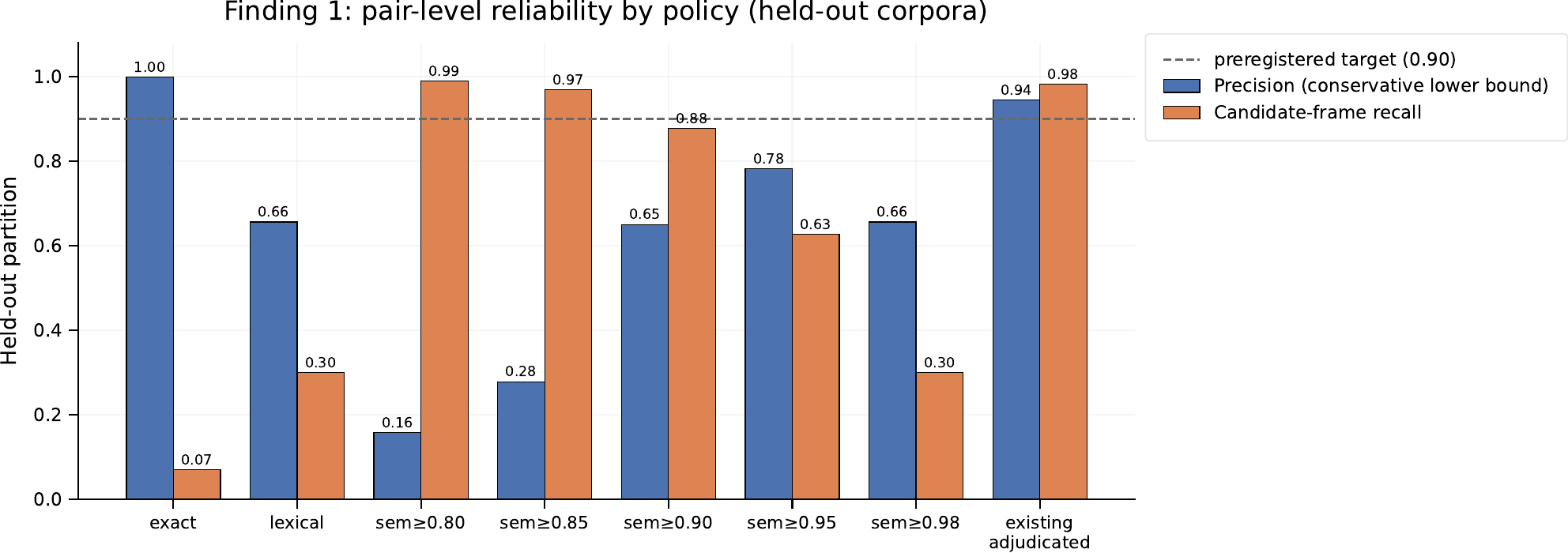}
\caption{Pair-level conservative precision and candidate-frame recall
for all eight policies, held-out partition only (three corpora and all
benchmark questions never used in calibration; full numbers, including
the calibration partition, in Table~\ref{tab:pairmetrics}). The dashed
line marks the preregistered 0.90 conservative-precision target
(\S\ref{sec:threshold}): only the existing adjudicated policy clears it
on this held-out partition while simultaneously retaining
candidate-frame recall above 0.90; its calibration conservative
precision is below the target.}
\label{fig:pairreliability}
\end{figure}

\begin{figure}[htbp]
\centering
\includegraphics[width=\linewidth]{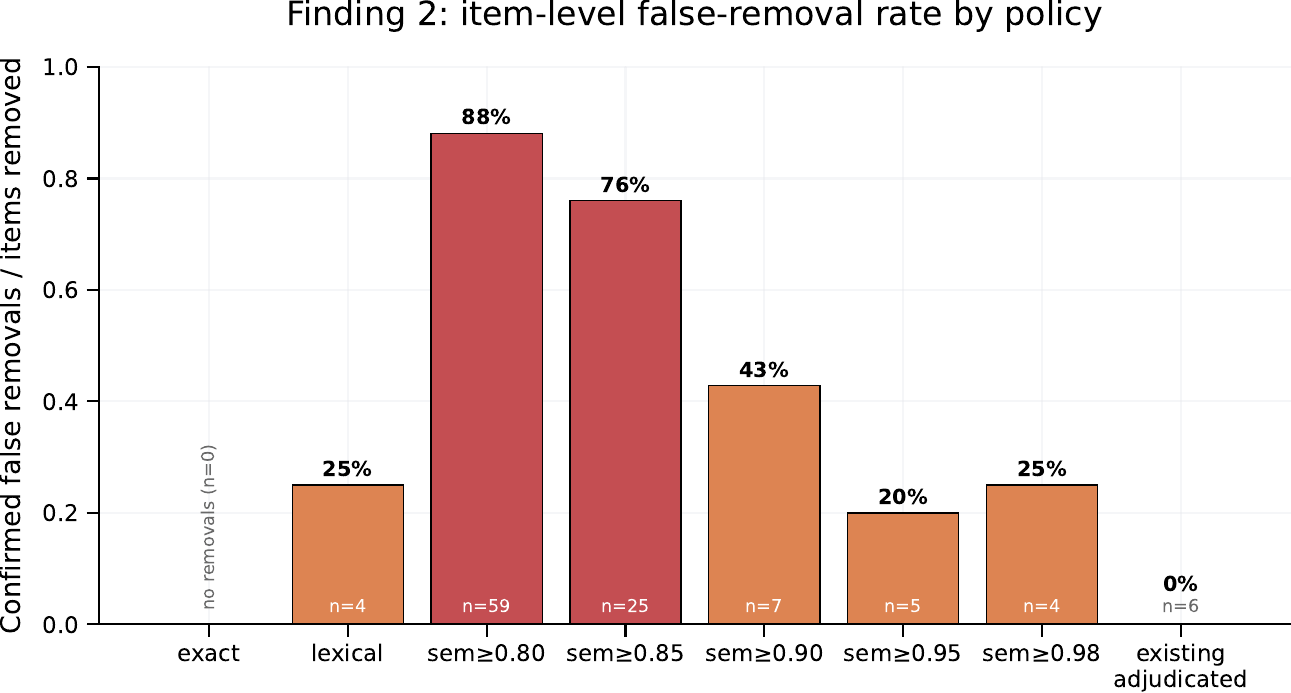}
\caption{Confirmed false-removal rate among sampled candidate-bearing
training items actually removed by each policy (complete-candidate-edge
item review; full counts in Table~\ref{tab:itemremoval}). ``exact''
removed none of the 80 sampled items and so has an undefined, not zero,
rate. The sampled semantic-policy false-removal rate generally falls
with a higher threshold, but rises from 20\% at 0.95 to 25\% at 0.98.
The existing adjudicated policy has zero confirmed false removals among
6 sampled removals, with one unresolved case; this does not establish
zero population-level risk.}
\label{fig:itemfalse}
\end{figure}

\begin{figure}[htbp]
\centering
\includegraphics[width=\linewidth]{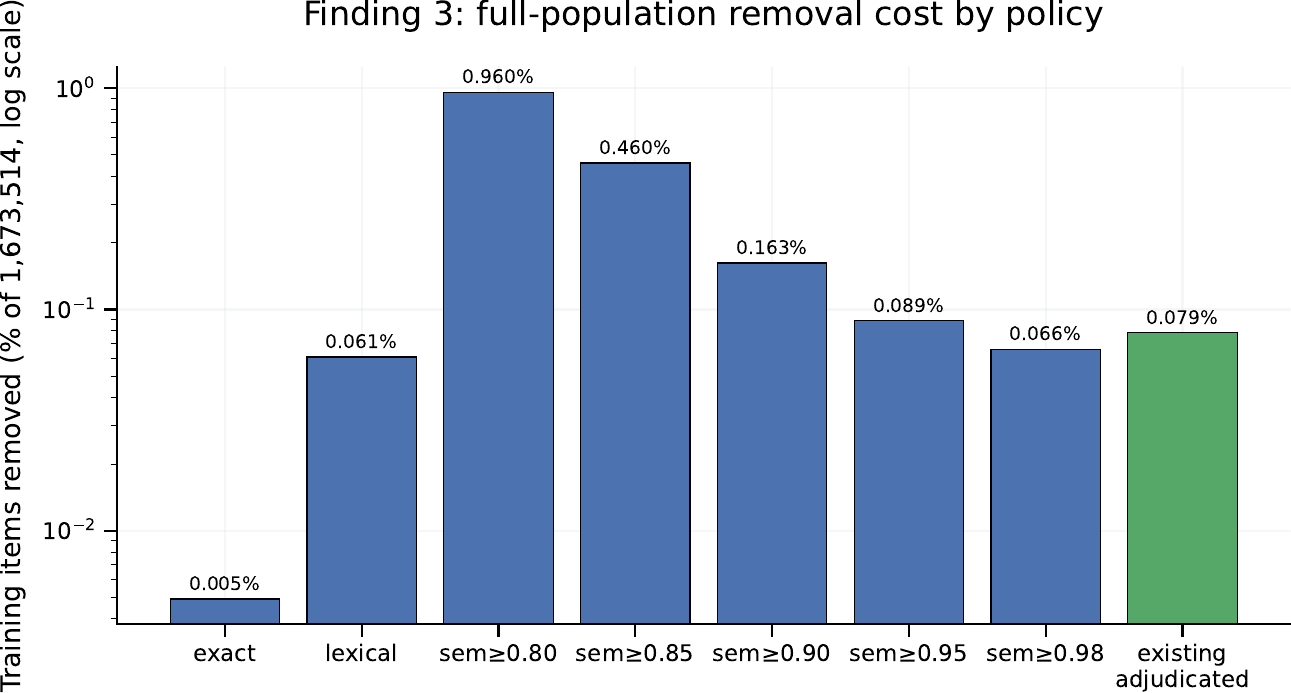}
\caption{Full-population removal cost, as a percentage of all
1{,}673{,}514 corpus-specific training items (not sampled; log-scaled
$y$-axis; full counts in Table~\ref{tab:coverage}). Threshold-0.80
semantic cleaning removes over twelve times as many items as the
existing adjudicated policy. Figure~\ref{fig:itemfalse} separately
reports false-removal rates among sampled removals, not demonstrated
population-level rates.}
\label{fig:coveragecost}
\end{figure}

Figures~\ref{fig:reviewsummary}--\ref{fig:coveragecost} recapitulate
the numbers in Tables~\ref{tab:pairmetrics}--\ref{tab:coverage} in visual
form, in the order the underlying evidence was collected: the
independent-review process itself (Figure~\ref{fig:reviewsummary}),
pair-level reliability on held-out corpora
(Figure~\ref{fig:pairreliability}), item-level confirmed false-removal
rates (Figure~\ref{fig:itemfalse}), and full-population removal cost
(Figure~\ref{fig:coveragecost}). Every value plotted is transcribed
directly from the corresponding table; the figures add no new
quantities and are provided purely to make the central
reliability-versus-recall trade-off (\S\ref{sec:result-semantic}) easier
to read at a glance.

\section{Discussion}
\label{sec:discussion}

\paragraph{More aggressive decontamination is not necessarily better.}
Our central result is a dissociation between recall and reliability
that a single overlap count cannot reveal. A detector that flags more
candidates is not automatically a better basis for a removal policy: in
our audited corpora, where many problems are built from recurring
templates (the same scenario with different numbers, renamed
quantities, or reordered constraints), a bare similarity threshold
cannot distinguish a changed-instance variant from genuine benchmark
exposure, and the gap is large---most permissive-threshold removals in
our sample were not contamination at all. This is a different question
from \emph{whether} contamination exists (already addressed by prior
work); it is a question about what a specific, deployable
cleaning rule actually does when applied.

\paragraph{Adjudication, not just a stronger embedding, is what closes
the gap.} The existing adjudicated policy in this pipeline is not a
higher embedding-similarity threshold; it is Tier-1 matches, plus an
LLM judge's case-by-case decision for ambiguous tier-2/tier-3
candidates. Our results suggest that this kind of explicit, per-pair
adjudication---not a better single scalar threshold---is what allows a
policy to simultaneously achieve high recall and high precision on data
it was not tuned on, within the held-out candidate frame evaluated here.
This has a direct practical implication for anyone
curating a new RLVR release: budget for adjudication of borderline
candidates, rather than only for a better retrieval model.

\paragraph{Item-level review changes the picture pair-level review alone
would give.} Had we reported only pair-level precision, a policy's
apparent conservatism could be inflated by under-reviewed items with a
second, unexamined genuine match. Reviewing every candidate edge for a
sampled item, rather than one pair at a time, is what lets us state the
0.80-threshold false-removal rate (88\% of removed sampled items) with
confidence that we are not conflating ``this particular pair looked
wrong'' with ``this item was wrongly removed'' within the reviewed
candidate frame. This descriptive sample rate is not a demonstrated
population-level false-removal rate.

\section{Limitations}
\label{sec:limitations}

\textbf{Candidate-frame-only recall.} Every recall estimate in this
paper is conditional on our existing three-tier candidate pipeline
having surfaced the pair at all; we cannot measure recall against
contamination that no tier's candidate generation step ever proposed,
and we do not claim whole-corpus recall. \textbf{Small item-level
sample.} Five candidate-bearing items per corpus (80 total) is
informative about false removals in the reviewed sample but gives wide
uncertainty for individual policies and corpora. In particular, zero
confirmed false removals among six adjudicated-policy removals, including
one unresolved case, does not establish zero false-removal risk. The
exploratory within-corpus bootstrap intervals we report are not a
substitute for a full stratified survey-design analysis, and should not
be read as precise population-level false-removal rates; the descriptive
52/59 (88\%) sample rate is not a demonstrated population rate.
\textbf{Two-reviewer, single-site annotation.} Both annotation passes
used the same two reviewers; we report raw agreement (89.4\% and
95.9\%) rather than a chance-corrected statistic, and do not have an
independent third reviewer pool to validate our own adjudication
decisions. Raw agreement can overstate reliability relative to
chance-corrected measures, and treating any single review pass as an
infallible ground truth is a known risk in annotation-quality research
\citep{james2026consensus,lavitas2021quality}; our reference labels
should be read as independently and blindly reviewed, not as an
unimpeachable gold standard. \textbf{Fixed corpus and benchmark
snapshots.} All counts are computed against specific downloaded
snapshots of sixteen corpora and six benchmarks; corpora that are
revised, or benchmarks released after our snapshot, are out of scope.
\textbf{No downstream training claim.} This paper evaluates
decontamination \emph{policies} against reviewed ground truth. It does
not train any model and makes no claim about whether applying any of
these policies changes downstream RLVR training outcomes; that causal
question is a distinct undertaking left to separate work.
\textbf{Math-only, English-only.} All audited corpora and benchmarks are
English-language competition or standardized-test mathematics; whether
our reliability ordering across policies transfers to other domains or
languages is untested.

\section{Conclusion}
\label{sec:conclusion}

We evaluated eight concrete decontamination policies for RLVR math
corpora against an independently, blindly dual-reviewed reference set
built specifically for this purpose, including a held-out partition of
corpora and benchmark questions never used for calibration and a
complete-candidate-edge item-level review designed to measure actual
false removal rather than single-pair precision. We find that permissive
semantic-similarity cleaning has high sampled false-removal costs,
although the sampled rates do not improve monotonically with a higher
threshold. No semantic threshold on our preregistered grid meets the
preregistered conservative-precision target on calibration. An existing
adjudicated policy already in use in this pipeline offers a more
favorable balance of resolved precision and candidate-frame recall on
the held-out corpora than the semantic-threshold alternatives tested,
at substantially lower full-corpus removal cost than the most permissive
policy. These results do not establish whole-corpus recall,
population-level false-removal rates, or zero false-removal risk.
We release the reviewed labels, the
per-policy removal decisions, and the full coverage accounting so that
future RLVR corpus releases can report and justify a specific
decontamination policy against measured reliability and cost, rather
than a single unqualified overlap count.

\bibliography{references}

\appendix
\section{Reproducibility Details}
\label{sec:appendix}

Table~\ref{tab:config} reports the full matching, embedding, and
annotation-design configuration used to generate candidates and draw the
reviewed samples, including the existing pipeline's fixed tier
thresholds (reused, not tuned by us) and the new study's preregistered
sampling and threshold-selection parameters.
\begin{table}[htbp]
\centering
\small
\setlength{\tabcolsep}{5pt}
\begin{tabular}{@{}p{0.46\linewidth}p{0.46\linewidth}@{}}
\toprule
Setting & Value \\
\midrule
\multicolumn{2}{@{}l}{\textit{Existing candidate pipeline (reused, not tuned by us)}} \\
Tier 2 long / short $n$-gram & 13 / 8 tokens \\
Tier 2 keep threshold (13-gram cov.\ / 8-gram cov.\ / Jaccard) & 0.20 / 0.30 / 0.40 \\
MinHash permutations & 128 \\
Embedding model & Qwen3-Embedding-0.6B \\
Embedding max sequence length & 256 tokens \\
Semantic candidate threshold (Tier 3 entry) & cosine $\geq 0.80$ \\
Existing adjudicated policy, lexical override & Jaccard $\geq 0.80$, or 13-gram cov.\ $\geq 0.80$ \\
Existing adjudicated policy, semantic override & cosine $\geq 0.95$ with matching final answer \\
\addlinespace
\multicolumn{2}{@{}l}{\textit{New study design (this paper)}} \\
Random seed & 20261010 \\
Pairs sampled per stratum & 2 \\
Benchmark-question calibration / held-out split & 60\% / 40\% \\
Corpora assigned entirely to held-out & bigmath\_rl, goosereason\_math, nemotron\_rl\_math\_v2 \\
Semantic threshold grid (preregistered) & 0.80, 0.85, 0.90, 0.95, 0.98 \\
Conservative-precision target (preregistered) & $\geq 0.90$ on calibration only \\
Items sampled per corpus (item-level review) & 5 \\
Reviewers per pair & 2, independent and blind \\
Adjudication & independent third pass, disagreements only \\
\bottomrule
\end{tabular}
\caption{Full configuration for candidate generation (reused from the
existing pipeline, listed here for reproducibility, not as a
contribution of this paper) and for the new independent-review study
design introduced in this paper. All study-design parameters were fixed
before any new label was reviewed. ``Pairs sampled per stratum'' refers
to the corpus $\times$ benchmark $\times$ (policy-flag, semantic-score
band) stratification described in \S\ref{sec:labels}.}
\label{tab:config}
\end{table}

\end{document}